\pdfoutput=1

\documentclass[11pt]{article}

\usepackage{acl}

\usepackage{times}
\usepackage{latexsym}

\usepackage[T1]{fontenc}

\usepackage[utf8]{inputenc}

\usepackage{microtype}

\usepackage{inconsolata}

\usepackage{graphicx}

\usepackage{amsmath, amssymb, amsthm}
\usepackage{multirow}
\usepackage{array}
\newcommand{\PreserveBackslash}[1]{\let\temp=\\#1\let\\=\temp}
\newcolumntype{C}[1]{>{\PreserveBackslash\centering}p{#1}}
\usepackage{xspace}  

\newcommand{\tool}{\textsc{LuxEmbedder}\xspace}
\newcommand{\benchmark}{\textsc{ParaLux}\xspace}
\newcommand{\paralleldata}{\textsc{LuxAlign}\xspace}

\title{\tool: A Cross-Lingual Approach to Enhanced Luxembourgish Sentence Embeddings}

\author{
 \textbf{Fred Philippy\textsuperscript{1,2}},
 \textbf{Siwen Guo\textsuperscript{1}},
 \textbf{Jacques Klein\textsuperscript{2}},
 \textbf{Tegawendé F. Bissyandé\textsuperscript{2}}
\\
\\
 \textsuperscript{1}Zortify S.A., Luxembourg \hfill \textsuperscript{2}University of Luxembourg, Luxembourg 
 \\
 \\
 \texttt{\{fred, siwen\}@zortify.com}
 \\
 \texttt{\{jacques.klein, tegawende.bissyande\}@uni.lu}
}

\begin{document}
\maketitle

\begin{abstract}
Sentence embedding models play a key role in various Natural Language Processing tasks, such as in Topic Modeling, Document Clustering and Recommendation Systems. However, these models rely heavily on parallel data, which can be scarce for many low-resource languages, including Luxembourgish. This scarcity results in suboptimal performance of monolingual and cross-lingual sentence embedding models for these languages. To address this issue, we compile a relatively small but high-quality human-generated cross-lingual parallel dataset to train \tool, an enhanced sentence embedding model for Luxembourgish with strong cross-lingual capabilities. Additionally, we present evidence suggesting that including low-resource languages in parallel training datasets can be more advantageous for other low-resource languages than relying solely on high-resource language pairs. Furthermore, recognizing the lack of sentence embedding benchmarks for low-resource languages, we create a paraphrase detection benchmark specifically for Luxembourgish, aiming to partially fill this gap and promote further research.\footnote{\url{https://github.com/fredxlpy/LuxEmbedder}}
\end{abstract}
\section{Introduction}
The development of sentence embedding models has been instrumental in applications such as Bitext Mining \citep{artetxe_massively_2019}, Information Retrieval \citep{thakur_beir_2021}, and most recently Retrieval Augmented Generation \citep{lewis_retrieval-augmented_2020}. Generative Large Language Models are not capable of handling these tasks as effectively, making sentence embedding models crucial in these areas. However, these models depend on large-scale parallel data to function effectively, a resource readily available for high-resource languages but sorely lacking for low-resource languages \citep{zhou_massively_2018}.

One way to address this issue is to apply cross-lingual sentence embedding models \citep{chidambaram_learning_2019, artetxe_massively_2019, reimers_making_2020, yang_multilingual_2020, feng_language-agnostic_2022,wang_english_2022}, which aim to embed various languages into a common shared representation space. This approach is intended to boost the performance of low-resource languages by leveraging cross-lingual transfer, where knowledge gained from high-resource languages contributes to the understanding and processing of low-resource languages. However, due to the significant differences in data availability, these models still exhibit a large performance gap between high-resource and low-resource languages.

Luxembourgish, a West-Germanic language spoken by about 400\,000 people, is one of the many languages that face this challenge. While translation models for Luxembourgish exist \citep{nllb_team_no_2022, song_letz_2023}, their performance remains significantly inferior to that of high-resource languages, hindering the creation of parallel data using methods like back-translation. This limitation also applies to general-purpose generative LLMs, making the direct creation of synthetic parallel data impractical as well. Our research aims to address this issue by collecting a comprehensive set of high-quality human-generated cross-lingual parallel data specifically for Luxembourgish. With this data, we train a sentence embedding model, \tool, tailored specifically for Luxembourgish by leveraging cross-lingual transfer.

Although cross-lingual sentence embedding models harness the strength of cross-lingual transfer to improve low-resource language performance, we argue that this does not eliminate the necessity for parallel data in these languages. Our findings demonstrate that incorporating these languages in parallel training datasets is essential, as it significantly improves alignment within cross-lingual models, particularly among other low-resource languages, in contrast to relying solely on high-resource language parallel data.

Another major challenge is the evaluation of sentence embedding models in low-resource languages, given that the primary benchmarks, such as MTEB \citep{muennighoff-etal-2023-mteb} and BEIR \citep{thakur_beir_2021}, predominantly support English and a few other high-resource languages.
To address this, we establish a new paraphrase detection benchmark for Luxembourgish, facilitating future research and improving the language’s representation in NLP. To thoroughly evaluate our enhanced model, \tool, we use our own benchmark along with three additional evaluation tasks. The results indicate that \tool outperforms not only other open-source models but also proprietary models in the majority of cases.

\section{Dataset \& Benchmark Construction}

\begin{figure*}[t]
    \centering
    \includegraphics[width=\textwidth]{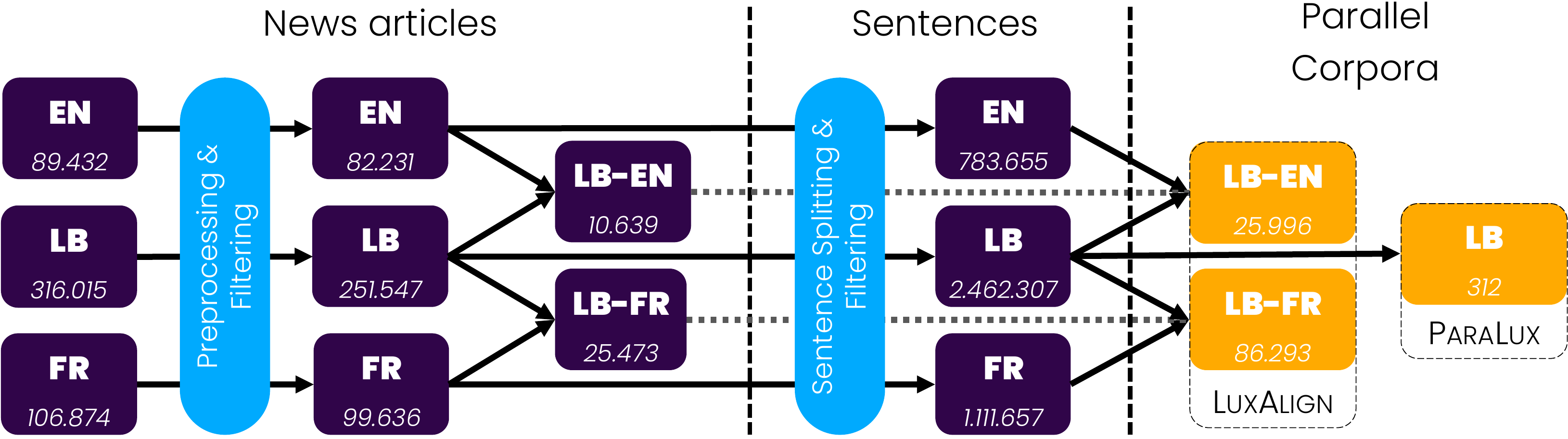}
    \caption{Our data construction workflow involves preprocessing and filtering news articles in English (\texttt{EN}), Luxembourgish (\texttt{LB}), and French (\texttt{FR}), aligning them through sentence embeddings, extracting parallel sentences from aligned article pairs to create \paralleldata, and generating the Luxembourgish paraphrase detection benchmark \benchmark. The numbers in \textit{italics} represent the number of documents used at each stage.}
    \label{fig:data_creation}
\end{figure*}

We create cross-lingual parallel data and a Luxembourgish paraphrase detection benchmark. See Appendix \ref{app:data_collection_processing} for details and Figure \ref{fig:data_creation} for an overview.

\subsection{Cross-Lingual Parallel Data (\paralleldata)} \label{sec:cl_data_collection_process}
We collect news articles from RTL.lu, a Luxembourgish news platform that publishes in Luxembourgish (LB), English (EN), and French (FR). Due to the lack of explicit mapping between language versions, we use the OpenAI text embedding model \texttt{text-embedding-3-small}\footnote{\url{https://platform.openai.com/docs/guides/embeddings/embedding-models}} to align articles across language pairs. LaBSE \citep{feng_language-agnostic_2022} is then employed to extract parallel sentences from these aligned pairs for LB-FR and LB-EN. 

\subsection{Luxembourgish Paraphrase Detection (\benchmark) Benchmark}
Then, we repeat the same process but focusing exclusively on Luxembourgish articles. Within each article, using the same setup, we extract parallel sentences, which can be considered near-paraphrases, from which we hand-pick high-quality samples for our benchmark. From these paraphrased pairs, we prompt \texttt{GPT-4o}\footnote{\url{https://openai.com/index/hello-gpt-4o/}} to generate adversarial negative samples for each pair. Given its limited language capabilities in Luxembourgish, the generated adversarial negative samples are then checked and, if needed, corrected by a human annotator to ensure high quality and accuracy. 

Through this methodology, we gather 25\,996 LB-EN, 86\,293 LB-FR samples for \paralleldata, and 312 samples for \benchmark.

\section{\tool}

\subsection{Training}
Given its cross-lingual capabilities and its already existing support of Luxembourgish, we use LaBSE \citep{feng_language-agnostic_2022} as our base model, which we further train on both LB-EN \& LB-FR parallel subsets from \paralleldata.

We train the model using a batch size of 16 for 3 epochs with a constant learning rate of $1 \times 10^{-6}$ using a contrastive loss function. We reserve 1\% of the data for evaluation, on which we evaluated every 500 steps, and retained the model with the best loss on the development set. The negative pairs for the loss function are created by randomly pairing each Luxembourgish sentence with the translation of another sentence from the dataset.

\subsection{Evaluation}
We comprehensively compare \tool's performance across multiple tasks against a variety of open-source and proprietary baseline models.

\subsection{Baselines}
We provide more details on the used models in Appendix \ref{app:baseline_models}.

\paragraph{Proprietary Models}
Developed by Cohere, \texttt{embed-multilingual-light-v3.0} and \texttt{embed-multilingual-v3.0} are multilingual embedding models, designed to handle over 100 languages, including Luxembourgish, producing embeddings of size 384 and 1\,024, respectively.

OpenAI's \texttt{text-embedding-3-small} and \texttt{text-embedding-3-large} models generate embeddings with dimensions of 1\,536 and 3\,072, respectively. Despite the native API feature for embedding shortening, we use the full dimensions in our experiments. While these models have been assessed on the multilingual MIRACL benchmark \citep{zhang-etal-2023-miracl}, there is no  official information on the number of supported languages.

\paragraph{Open-Source Models} We also compare \tool against two open-source multilingual sentence embedding models that support Luxembourgish. These models are LaBSE \citep{feng_language-agnostic_2022}, which generates cross-lingual sentence embeddings for 109 languages, and LASER \citep{artetxe_massively_2019, heffernan_bitext_2022}, which incorporates a multilingual teacher sentence embedding model and language-specific student models for 200 languages.

We further extend our evaluation to include mBERT, a multilingual BERT \citep{devlin_bert_2019}, variant pre-trained on 104 languages, and LuxemBERT \citep{lothritz_luxembert_2022}, a monolingual Luxembourgish BERT model. In our experiments, we leverage both CLS embeddings and MEAN-pooled embeddings from these models.

\subsection{Evaluation Tasks}
Additional details on the specific evaluation setup can be found in Appendix \ref{app:evaluation_tasks}.

\paragraph{Zero-Shot Classification} \
Using SIB-200 \citep{adelani_sib-200_2024}, a 7-class classification dataset, we perform similarity-based zero-shot classification. First, we fill each label into a pre-defined template sentence, and separately encode both the input document and all potential template-embedded labels. Then, the class with the most similar embedding to the input document is chosen, assessing the model’s ability to generalize to new, unseen tasks without any task-specific training. To account for variability, we repeat this process for 5 different label templates and report the average performance.

\paragraph{Cross-Lingual Transfer} \
For cross-lingual transfer performance, we use the embeddings generated by the respective model to fine-tune a classifier on the SIB-200 dataset in six different high-resource source languages and evaluate directly on the Luxembourgish test set. 

\paragraph{Bitext Mining} \
We evaluate the model's proficiency in accurately retrieving or matching parallel sentence pairs from a bilingual corpus using the Tatoeba dataset. Since the original Tatoeba test set \citep{artetxe_massively_2019} does not include Luxembourgish, we use the LB-EN, LB-NL, and LB-DE test sets developed by the \textit{Tatoeba Translation Challenge} \citep{tiedemann_tatoeba_2020}.

\paragraph{\benchmark} \
Lastly, we evaluate the model on our newly created benchmark for paraphrase detection. This task involves determining which of two sentences is a paraphrase of a given anchor sentence. It tests the model’s ability to discern nuanced semantic equivalence, which is critical for applications like plagiarism detection, question answering, and information retrieval.

\subsection{Results} \label{sec:results}
\tool demonstrates superior performance among open-source models in all four tasks and even outperforms all tested proprietary models in 3 out of 4 tasks (Table \ref{tab:Luxemebdder_results}). Only \texttt{text-embedding-3-large} model shows superior cross-lingual transfer performance.

In particular, we observe considerable improvements in \tool’s performance on both monolingual tasks, Zero-Shot Classification and Paraphrase Detection, relative to its base model, LaBSE. This confirms the efficacy of our cross-lingual approach for Luxembourgish.

\begin{table*}[htbp!]
\centering
\renewcommand{\arraystretch}{1.2}
\begin{tabular}{c|c|C{1.6cm}|C{1.6cm}|C{1.6cm}|C{1.6cm}}
 & \textbf{Model} & \begin{tabular}[c]{@{}c@{}}\textbf{CL}\\ \textbf{Transfer}\end{tabular} & \begin{tabular}[c]{@{}c@{}}\textbf{Bitext}\\ \textbf{Mining}\end{tabular} & \begin{tabular}[c]{@{}c@{}}\textbf{Zero-Shot}\\ \textbf{Classific.}\end{tabular}  & \textbf{\benchmark} \\ \hline
\multirow{4}{*}{\rotatebox[origin=c]{90}{Proprietary}} & \textbf{Cohere/embed-multilingual-light-v3.0} & 70.89 & 50.10 & 40.39 & 37.50 \\
 & \textbf{Cohere/embed-multilingual-v3.0} & 79.49 & 59.38 & 53.33 & 49.04 \\
 & \textbf{OpenAI/text-embedding-3-small} & 72.59 & 39.30 & 40.20 & 15.71 \\
 & \textbf{OpenAI/text-embedding-3-large} & \textbf{86.25} & 56.04 & 58.82 & 26.28 \\ \hline
\multirow{7}{*}{\rotatebox[origin=c]{90}{Open-Source}} & \textbf{mBERT (MEAN)} & 70.53 & 28.44 & 15.49 & 5.13 \\
 & \textbf{mBERT(CLS)} & 70.20 & 22.27 & 13.73 & 4.81 \\
 & \textbf{LuxemBERT (MEAN)} & 48.47 & 30.33 & 14.02 & 7.69 \\
 & \textbf{LuxemBERT(CLS)} & 56.86 & 21.94 & 33.73 & 14.42 \\
 & \textbf{LASER} & 62.70 & 62.96 & 11.08 & 16.03 \\
 & \textbf{LaBSE} & 80.88 & 70.11 & 43.24 & 38.14 \\
 & \textbf{\tool} & \underline{83.39} & \underline{\textbf{70.24}} & \underline{\textbf{65.59}} & \underline{\textbf{52.24}} \\ \hline
\end{tabular}
\caption{Comparison of LuxEmbedder with various open-source and proprietary models across two cross-lingual and two monolingual tasks. We report accuracy for all 4 tasks. The best overall performance for each task is highlighted in \textbf{bold}, while the best performance among open-source models is \underline{underlined}.}
\label{tab:Luxemebdder_results}
\end{table*}

\section{Cross-Lingual Alignment} \label{sec: crosslingual_alignment}
In this section, we investigate the impact of fine-tuning models on parallel data for cross-lingual alignment between and within high-resource (HR) and low-resource (LR) languages. 

\paragraph{Experimental Setup}
To measure the cross-lingual alignment, we use Flores-200 \citep{nllb_team_no_2022}, which includes parallel sentences across 200 languages, making it an ideal resource for assessing cross-lingual alignment. We use the Centered Kernel Alignment (CKA) method \citep{kornblith_similarity_2019} to calculate the level of alignment by comparing the embeddings of parallel sentences from different languages.

We fine-tune LaBSE on three different language pairs: LB-EN, LB-FR, and EN-FR\footnote{Created using the same process as described in \S \ref{sec:cl_data_collection_process}.}, each time using 20\,000 parallel sentences from our newly compiled datasets. After fine-tuning, we assess cross-lingual alignment by comparing alignment \underline{within} HR languages and LR languages, as well as \underline{between} LR and HR languages\footnote{As HR and LR languages we select the 10 languages with the most and least training data in LaBSE which are also covered by Flores-200.}.

\begin{figure}[h!]
    \centering
    \includegraphics[width=\linewidth]{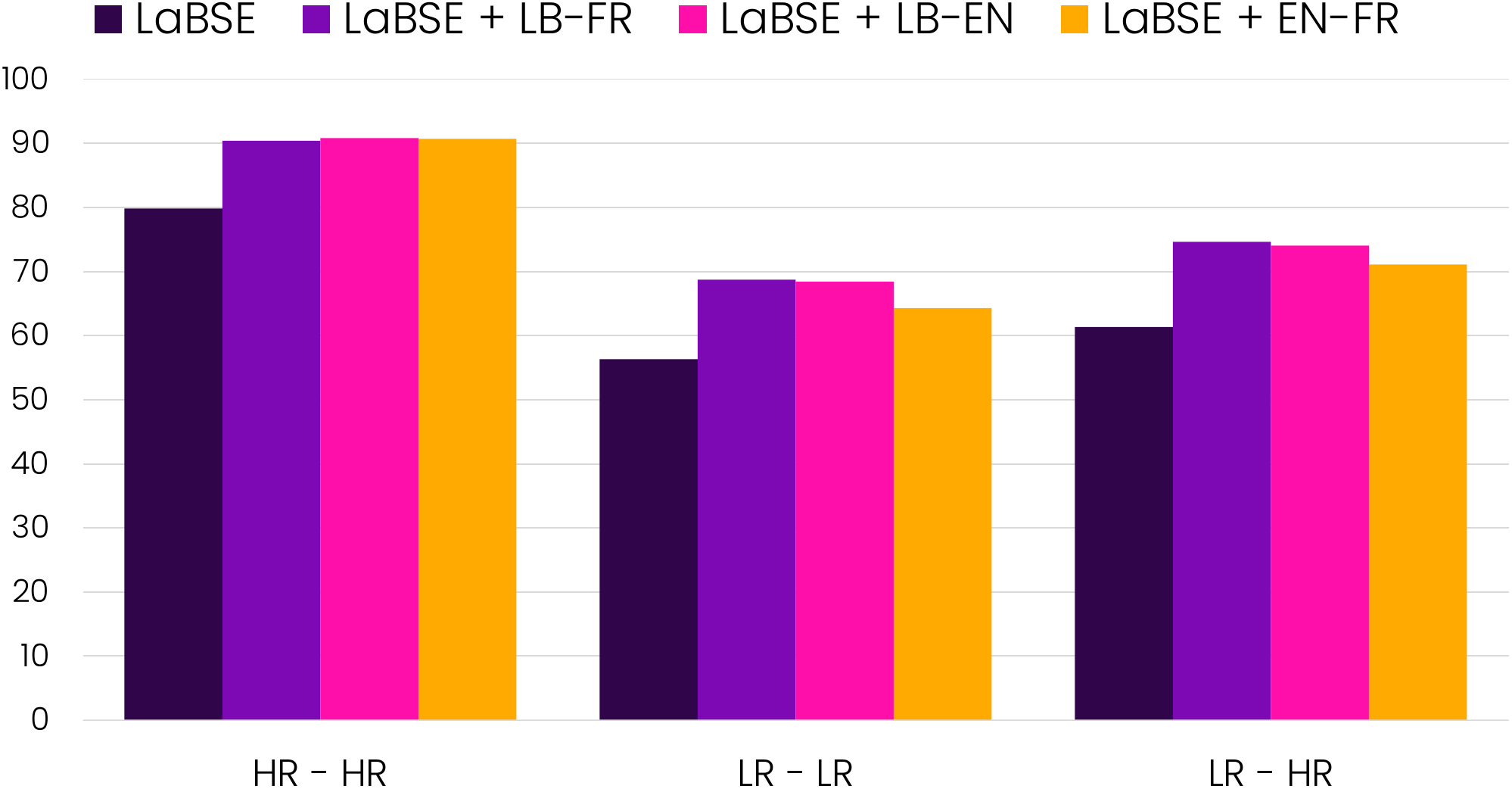}
    \caption{The alignment of language-specific embedding spaces within and between high-resource (HR) and low-resource (LR) languages, measured using the CKA method, is shown for LaBSE before and after fine-tuning on LB-EN, LB-FR, and EN-FR parallel data. The exact values for all language pairs are provided in Figure \ref{fig:cka_heatmaps}.}
    \label{fig:bar_chart_alignment}
\end{figure}

\paragraph{Results}
Our observations (Figure \ref{fig:bar_chart_alignment}) reveal that when fine-tuning on parallel data, the alignment within the model generally increases. HR languages benefit equally from fine-tuning on any of the three language pairs. However, we observe that the alignment of LR languages benefits more when Luxembourgish is part of the training data compared to fine-tuning on HR language pairs alone.

These results indicate the critical importance of including LR languages, such as Luxembourgish, when collecting parallel data. Incorporating LR in the training process enhances cross-lingual alignment, not only for the respective language pair but also for other LR languages, more effectively than focusing solely on HR languages.
\section{Conclusion}
Sentence embedding models struggle with low-resource languages due to a shortage of parallel data. To address this problem, we collected high-quality, human-generated cross-lingual parallel data for Luxembourgish and developed an enhanced version of a cross-lingual sentence embedding model specifically adapted to Luxembourgish. This model outperforms open-source as well as proprietary models in almost all evaluations conducted in our study. Our findings also stress the importance of incorporating low-resource languages in parallel data collection, as evidence suggests that this enhances embedding alignment for both the target language and other low-resource languages within the same model more effectively than using high-resource language pairs alone. Therefore, we believe this research encourages further creation of parallel corpora for low-resource languages.
\section*{Limitations}
It is important to note that we do not compare our embedding model against general-purpose generative LLMs. We acknowledge that some of these models, which are significantly larger in terms of parameter count, may outperform \tool in certain tasks. Nonetheless, the primary objective of our paper is not to compete with generative models . Instead, our focus is on providing a robust sentence embedding model capable of solving specific tasks such as information retrieval, document clustering, and similar applications where generative language models may not be as effective.

Additionally, we acknowledge that our data is limited to the news domain, due to its availability. However, our goal is to use this data to boost the model’s retrieval performance, facilitating future expansion into various other domains by mining a more diverse range of parallel data.
\section*{Ethical Statement}
In the newly created \benchmark benchmark, the adversarial counterparts of the paraphrases have been edited in a way that some of the edited sentences may contain non-factual information. Therefore, we strongly recommend using this data solely, as designed, for evaluation purposes and not for training, to ensure the integrity of model development.

Furthermore, our datasets, based on news articles, naturally include the names of individuals. As the text is publicly available and anonymization would greatly diminish data quality, we chose not to anonymize it.  We believe that preserving the original context of publicly accessible information is essential for maintaining data integrity and the effectiveness of our research.
\section*{Acknowledgments}
We are grateful to RTL Luxembourg for providing the raw data necessary for our research. Their support significantly facilitated our efforts.


\bibliography{anthology, zotero}

\appendix
\section{Data Collection \& Processing} \label{app:data_collection_processing}
Here, we outline the method used to create cross-lingual training data and the paraphrase detection benchmark, providing examples in Tables \ref{tab:cl_dataset_examples} and \ref{tab:lb_dataset_examples}. 


\paragraph{Processing Articles}
We gather news articles from the Luxembourgish news platform RTL\footnote{\url{https://www.rtl.lu}} written in Luxembourgish, French, and English, covering different time periods: from January 1, 1999 for Luxembourgish, from September 1, 2011 for French, and from January 1, 2018 for English, up until May 10, 2024. We first remove all URL tags and extraneous metadata, and filter out articles with fewer than 100 characters, as these are often just traffic or sports updates, which were not relevant for our study. To ensure linguistic accuracy, we use the OpenLID \citep{burchell_open_2023} to identify and exclude articles that are not in the intended language.

\paragraph{Article Matching}
Subsequently, we embed each article using the OpenAI \texttt{text-embedding-3-small} model to facilitate cross-language article matching. To identify potential parallel articles in different languages, we first narrow down the candidates by considering only those articles published within a one-day window of the target article. Among these candidates, we select the one with the highest cosine similarity to the target article’s embedding, provided the similarity score exceeds 0.65. 

\paragraph{Sentence Extraction}
In parallel, we extract sentences from each article using the NLTK\footnote{\url{https://www.nltk.org}} library. For Luxembourgish, in the absence of a dedicated sentence tokenizer, we use the German tokenizer. After splitting the articles into sentences, we employ OpenLID once again to remove any sentences identified as being in the wrong language. Additionally, we filter out sentences with fewer than 10 characters or fewer than three words.

\paragraph{Sentence Matching}
Next, we embed each sentence using LaBSE, focusing on sentences from articles already matched with articles in another language. For each sentence, we restrict the candidates to sentences from the corresponding matched article, minimizing the risk of false positives. We then select the candidate sentence with the highest cosine similarity, provided it exceeds a similarity threshold of 0.7. After identifying all sentence pairs, we filter out pairs where the length difference is greater than 50\%. To create a seed dataset for \benchmark, we replicate this process within Luxembourgish articles alone.

\section{Training and Evaluation Details for \tool}
All our training processes and experiments were run on 4 A100 GPUs within a few hours.

\subsection{Training}
Given a sentence embedding model $\mathcal{M}_\theta$ with parameters $\theta$, for a sentence pair $(x_1, x_2)$ and its label $y$ (1 if positive pair, 0 if negative pair), the contrastive loss function is defined as:

\begin{equation}
    \begin{aligned}
        \mathcal{L} & \left( \theta,  (x_1,x_2,y) \right) = \\ 
        & \frac{1}{2} \left[ y \cdot \text{D}^2 + (1 - y) \cdot \max(0, \text{m} - \text{D})^2 \right]
    \end{aligned}
\end{equation}

where
\begin{itemize}
\item $D = d\left(\mathcal{M}_\theta(x_1), \mathcal{M}_\theta(x_2)\right)$
\item $m$ is the margin value, defining the minimum distance that samples withing a negative pair should have
\end{itemize}

with $m=0.5$ and $d$ being the cosine distance in our experiments.

\subsection{Evaluation} \label{app:evaluation}

\subsubsection{Baseline Models} \label{app:baseline_models}
Due to the proprietary nature of Cohere’s models, \texttt{embed-multilingual-light-v3.0} and \texttt{embed-multilingual-v3.0}, as well as OpenAI’s \texttt{text-embedding-3-small} and \texttt{text-embedding-3-large}, detailed information about their training data and model architecture is not publicly available. We refer readers to their online documentation\footnote{\url{https://cohere.com/blog/introducing-embed-v3}} \footnote{\url{https://openai.com/index/new-embedding-models-and-api-updates/}} for any details.

Our experiments with open-source models involve base multilingual BERT (cased) \citep{devlin_bert_2019} and LuxemBERT \citep{lothritz_luxembert_2022}. These models feature identical architectures, including 12 attention heads and 12 transformer blocks, each with a hidden size of 768. mBERT’s vocabulary size is 30\,000, whereas LuxemBERT’s is 119\,547. Both models have about 110 million parameters.

Additionally, we incorporate LaBSE \citep{feng_language-agnostic_2022}, which also serves as the foundational model for \tool. LaBSE is derived from the base multilingual BERT (cased) but features an expanded vocabulary of 501\,153 tokens. It has been trained using a combination of monolingual data and bilingual translation pairs.

\subsubsection{Evaluation Tasks} \label{app:evaluation_tasks}
\paragraph{Cross-Lingual Transfer} \ \\
To assess cross-lingual transfer performance, we use embeddings from the respective model to fine-tune a classifier on the SIB-200 \citep{adelani_sib-200_2024} dataset in several high-resource source languages, then evaluate it directly on the Luxembourgish test set.

The SIB-200 dataset includes over 200 languages, with 701 training, 99 development and 204 test samples per language.

In our experiments, however, we only train separately on French, English, German, Japanese, Chinese, and Russian. Additionally, we fine-tune on Luxembourgish, but this is not included in the average performance reported in Table \ref{tab:Luxemebdder_results}. The classifier is a simple linear layer with 7 output nodes, trained with the Adam optimizer and the cross-entropy loss function. Training is performed for 500 epochs with a constant learning rate of $1e^{-2}$. We evaluate the classifier once per epoch and select the model with the best development loss. Each training process is repeated 4 times using different seeds to ensure robustness, and we report the average performance per source language in Table \ref{tab:CL_transfer_full_results}.

\paragraph{Zero-Shot Classification} \ \\
To assess the zero-shot classification capabilities of different model, we again use the SIB-200 dataset \citep{adelani_sib-200_2024}. We independently encode the input and all potential labels,  integrating each label within a prompt template. The class whose embedding has the highest cosine similarity to the input document is selected.

We use five different prompt templates to evaluate the classification performance and report the average performance per template in Table \ref{tab:ZSC_full_results}. These templates are: 
\begin{enumerate}
    \item \texttt{[LABEL]}
    \item \texttt{An dësem Beispill geet et em [LABEL].} \\
        \textit{This example is about [LABEL].}
    \item \texttt{D'Thema vun dësem Text ass [LABEL].} \\
        \textit{The topic of this text is [LABEL].}
    \item \texttt{Hei gëtt iwwer [LABEL] geschwat.} \\
    \textit{Here we are talking about [LABEL].}
    \item \texttt{Dëst Dokument beschäftegt sech mat [LABEL].} \\
    \textit{This document deals with [LABEL].}
\end{enumerate}

The labels in Luxembourgish we use in this classification task are \texttt{Technologie} (\textit{technology}), \texttt{Reesen} (\textit{travel}), \texttt{Politik} (\textit{politics}), \texttt{Gesondheet} (\textit{health}), \texttt{Ennerhalung} (\textit{entertainment}), \texttt{Geographie} (\textit{geography}) and \texttt{Sport} (\textit{sports}).

\paragraph{Bitext Mining} \ \\
We initially considered the Tatoeba dataset, but it lacks Luxembourgish in the original set. Instead, we used Luxembourgish-English, Luxembourgish-Dutch, and Luxembourgish-German test sets from the \textit{Tatoeba Translation Challenge} \citep{tiedemann_tatoeba_2020}, which include 346 LB-EN, 291 LB-EN, and 292 LB-DE sample pairs.\footnote{\url{https://huggingface.co/datasets/Helsinki-NLP/tatoeba_mt}} We conducted experiments in both retrieval directions and reported the full results in Table \ref{tab:tatoeba_full_results}.

\paragraph{\benchmark} \ \\
To assess performance on \benchmark, the model encoded the anchor sentence and both paraphrase candidates. The candidate with the greatest cosine similarity to the anchor was chosen as the predicted paraphrase.

\section{Full Results}
Here, we report the full experimental results from the evaluations on Cross-Lingual Transfer (Table \ref{tab:CL_transfer_full_results}), Zero-Shot Classification (Table \ref{tab:ZSC_full_results}) and Bitext Mining (Table \ref{tab:tatoeba_full_results}) conducted in Section \ref{sec:results}.

\section{Details on the Cross-Lingual Alignment Experiments}
In Section \ref{sec: crosslingual_alignment}, we measure the alignment of language-specific subspaces using the Centered Kernel Alignment (CKA) method \cite{kornblith_similarity_2019}. The CKA score of two representation matrices \( X \in \mathbb{R}^{N \times m} \) and \( Y \in \mathbb{R}^{N \times m} \), where \( N \) is the number of samples and \( m \) is the embedding dimension of the model, when using a linear kernel, is given by
$$
CKA(X, Y) = 1 - \frac{\|XY^T\|_F^2}{\|XX^T\|_F \|YY^T\|_F}
$$

where \(\|\cdot\|_F\) is the Frobenius norm. \\

Since parallel cross-lingual data is essential for computing the CKA across various languages, we use the Flores-200 dataset \citep{nllb_team_no_2022}, which includes human-curated translations between English and 204 other languages. Specifically, we use the devtest split, containing 1\,012 aligned sentences per language.

We choose the 10 languages with the highest and lowest amounts of training data in LaBSE, which are also included in Flores-200, to represent the HR and LR languages. As LR languages, we use \texttt{bod}, \texttt{snd}, \texttt{tuk}, \texttt{ydd}, \texttt{wol}, \texttt{asm}, \texttt{smo}, \texttt{xho}, \texttt{nya}, and \texttt{sot}. As HR languages, we use \texttt{eng}, \texttt{rus}, \texttt{jpn}, \texttt{zho}, \texttt{fra}, \texttt{deu}, \texttt{por}, \texttt{nld}, \texttt{spa}, and \texttt{pol}. 

\begin{figure*}
    \centering
    \includegraphics[width=1\linewidth]{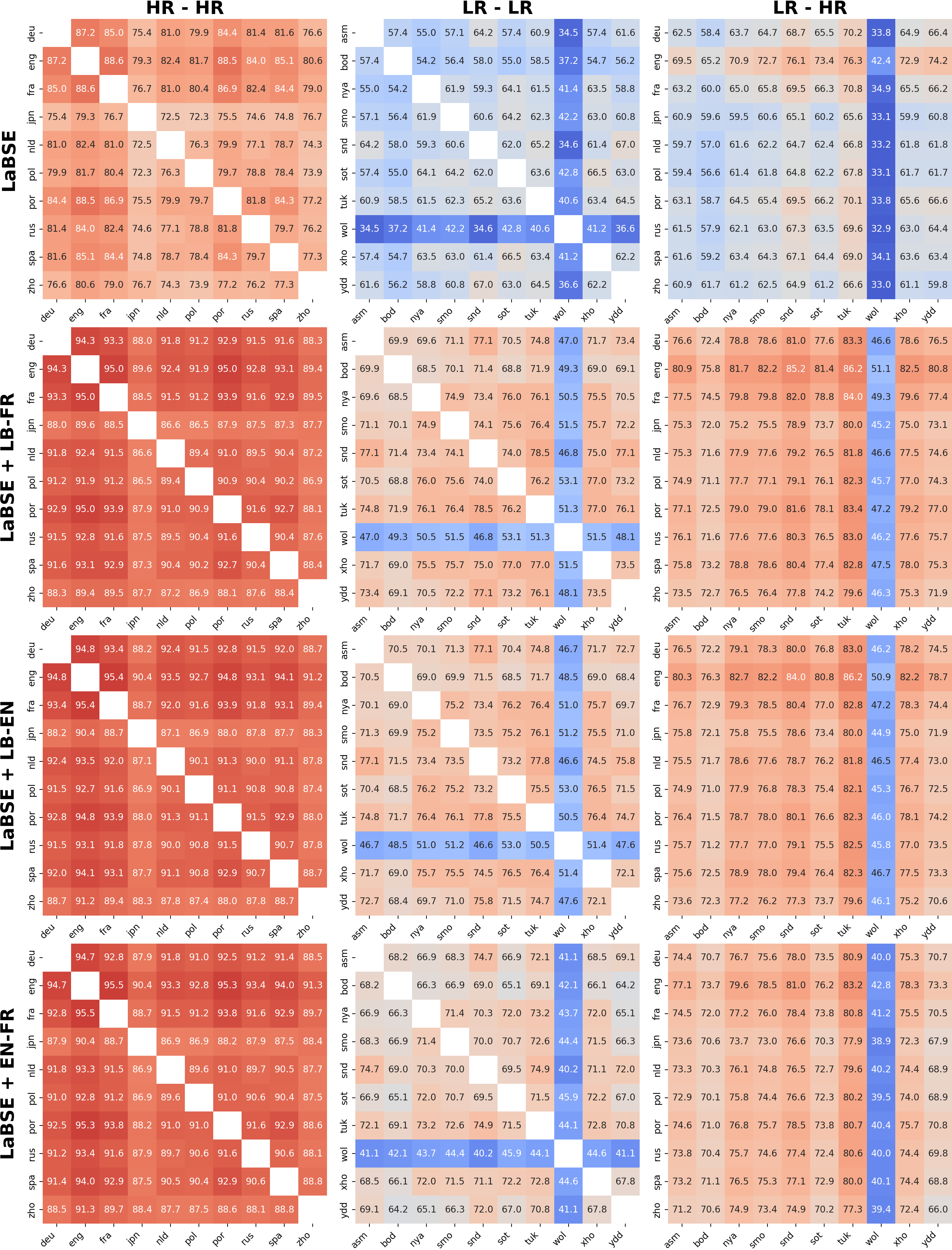}
    \caption{Alignment of language-specific embedding spaces within and between high-resource (HR) and low-resource (LR) languages, measured using the CKA method for LaBSE before and after fine-tuning on LB-EN, LB-FR, and EN-FR parallel data.}
    \label{fig:cka_heatmaps}
\end{figure*}

The exact CKA values across all language pairs are provided in Figure \ref{fig:cka_heatmaps}. \\

\begin{table*}[]
    \centering
    \renewcommand{\arraystretch}{1.5}
    \begin{tabular}{p{0.45\linewidth}|p{0.45\linewidth}}
         \multicolumn{1}{c}{\textbf{Luxembourgish Sentence}} & \multicolumn{1}{|c}{\textbf{English/French Sentence}} \\ \hline \hline
         D'Police sicht no engem Mann, deen an der Stad mat enger geklauter Kreditkaart Suen opgehuewen huet. & The police is looking for a man who withdrew money with a stolen credit card in Luxembourg City. \\ \hline
         D'Temperaturen am Grand-Duché sinn an der Moyenne em 1.3 Grad an d'Luucht gaangen. & Temperatures in the Grand Duchy have risen by 1.3 degrees on average. \\ \hline
         Déi Petitioun ass vun 336.000 Persounen aus 112 Länner ënnerschriwwe ginn. & Cette pétition a été signée par 336.000 personnes originaires de 112 pays. \\ \hline 
         Am September 2013 hat fir d'éischte Kéier e Lëtzebuerger den Jackpot gewonnen. & En septembre 2013, un Luxembourgeois avait pour la 1e fois remporté le jackpot. \\ \hline
    \end{tabular}
    \caption{Examples from the compiled parallel LB-EN \& LB-FR dataset \paralleldata.}
    \label{tab:cl_dataset_examples}
\end{table*}

\begin{table*}[]
    \centering
    \renewcommand{\arraystretch}{1.5}
    \begin{tabular}{p{0.31\linewidth}|p{0.31\linewidth}|p{0.31\linewidth}}
         \multicolumn{1}{c}{\textbf{Anchor Sentence}} & \multicolumn{1}{|c|}{\textbf{Paraphrase}} &  \multicolumn{1}{c}{\textbf{Not Paraphrase}} \\ \hline \hline
         Mexiko gewënnt 3-1 géint Kroatien. & Kroatien verléiert 1-3  géint Mexiko. & Kroatien gewënnt 3-1 géint Mexiko. \\ 
         \textit{\small Mexico wins 3-1 against Croatia.} & \textit{\small Croatia loses 3-1 against Mexico.} & \textit{\small Croatia wins 3-1 against Mexico.} \\ \hline
         De Sträit tëscht Süd- a Nordkorea spëtzt sech weider zou. & D'Verhältnis tëscht Nord- a Südkorea gëtt ëmmer méi schlecht. & De Sträit tëscht Süd- a Nordkorea entspaant sech weider. \\ 
         \textit{\small The dispute between South and North Korea continues to escalate.} & \textit{\small The relationship between South and North Korea is getting worse and worse.} & \textit{\small The dispute between South and North Korea continues to ease.} \\ \hline
    \end{tabular}
    \caption{Examples from the newly created Luxembourgish paraphrase detection benchmark \benchmark.}
    \label{tab:lb_dataset_examples}
\end{table*}

\begin{table*}[]
\centering
\renewcommand{\arraystretch}{1.2}
\setlength{\tabcolsep}{5pt}
\begin{tabular}{c|c|cccccc|c}
 & \multirow{2}{*}{\textbf{Model}} & \multicolumn{7}{c}{\textbf{Source Language}} \\ \cline{3-9} 
 &  & \textbf{de} & \textbf{en} & \textbf{fr} & \textbf{jp} & \textbf{ru} & \textbf{zh} & \textbf{lb} \\ \hline
\multirow{4}{*}{\rotatebox[origin=c]{90}{Proprietary}} & \textbf{Cohere/embed-multilingual-light-v3.0} & 69.24 & 70.10 & 69.00 & 73.28 & 74.26 & 69.49 & 76.96 \\
 & \textbf{Cohere/embed-multilingual-v3.0} & 79.90 & 82.60 & 76.47 & 79.29 & 78.19 & 80.51 & 83.09 \\
 & \textbf{OpenAI/text-embedding-3-small} & 76.96 & 73.53 & 68.14 & 72.18 & 74.02 & 70.71 & 77.08 \\
 & \textbf{OpenAI/text-embedding-3-large} & \textbf{87.75} & \textbf{84.44} & \textbf{85.54} & \textbf{85.54} & \textbf{87.75} & \textbf{86.52} & \textbf{88.36} \\ \hline
\multirow{7}{*}{\rotatebox[origin=c]{90}{Open-Source}} & \textbf{mBERT (MEAN)} & 72.67 & 70.96 & 73.28 & 65.44 & 72.55 & 68.26 & 75.12 \\
 & \textbf{mBERT (CLS)} & 70.10 & 68.75 & 71.08 & 67.89 & 73.28 & 70.10 & 75.25 \\
 & \textbf{LuxemBERT (MEAN)} & 79.41 & 74.02 & 81.74 & 23.41 & 9.56 & 22.67 & 82.48 \\
 & \textbf{LuxemBERT (CLS)} & 80.02 & 81.86 & 79.29 & 35.05 & 28.92 & 36.03 & \underline{84.44} \\
 & \textbf{LASER} & 62.01 & 61.76 & 63.36 & 61.76 & 62.38 & 64.95 & 57.84 \\
 & \textbf{LaBSE} & 80.51 & 79.90 & 80.88 & 82.35 & 80.39 & 81.25 & 80.15 \\
 & \textbf{\tool} & \underline{83.82} & \underline{82.48} & \underline{82.84} & \underline{83.58} & \underline{83.33} & \underline{84.31} & 84.31 \\ \hline
\end{tabular}
\caption{\textbf{Cross-lingual Transfer Performance}: Comparative results of models on the SIB-200 dataset. Linear classifiers were trained using model embeddings in various source languages and evaluated on the Luxembourgish test set. The table shows average performance from 4 experiment iterations. The best overall performance for each source language is highlighted in \textbf{bold} while the best performance among open-source models is \underline{underlined}.}
\label{tab:CL_transfer_full_results}
\end{table*}

\begin{table*}[]
\centering
\renewcommand{\arraystretch}{1.2}
\begin{tabular}{c|c|ccccc}
 & \multirow{2}{*}{\textbf{Model}} & \multicolumn{5}{c}{\textbf{Label Template}} \\ \cline{3-7} 
 &  & \textbf{1} & \textbf{2} & \textbf{3} & \textbf{4} & \textbf{5} \\ \hline
\multirow{4}{*}{Proprietary} & \textbf{Cohere/embed-multilingual-light-v3.0} & 42.65 & 42.65 & 35.29 & 42.16 & 39.22 \\
 & \textbf{Cohere/embed-multilingual-v3.0} & 45.59 & 56.86 & 53.92 & 58.82 & 51.47 \\
 & \textbf{OpenAI/text-embedding-3-small} & 24.02 & 46.08 & 35.29 & 48.04 & 47.06 \\
 & \textbf{OpenAI/text-embedding-3-large} & 42.65 & 65.20 & 67.16 & 57.84 & 61.27 \\ \hline
\multirow{7}{*}{Open-Source} & \textbf{mBERT (MEAN)} & 10.78 & 15.69 & 15.69 & 17.65 & 17.65 \\
 & \textbf{mBERT (CLS)} & 11.27 & 16.67 & 11.76 & 16.18 & 12.75 \\
 & \textbf{LuxemBERT (MEAN)} & 9.31 & 9.31 & 14.71 & 15.69 & 21.08 \\
 & \textbf{Luxembert (CLS)} & 9.31 & 41.67 & 23.53 & 50.49 & 43.63 \\
 & \textbf{LASER} & 13.73 & 10.78 & 10.78 & 10.29 & 9.80 \\
 & \textbf{LaBSE} & 38.24 & 45.10 & 44.12 & 47.55 & 41.18 \\
 & \textbf{\tool} & \textbf{55.88} & \textbf{68.14} & \textbf{68.14} & \textbf{69.61} & \textbf{66.18} \\ \hline
\end{tabular}
\caption{\textbf{Zero-Shot Classification Performance} on the SIB-200 datasets for five different label templates. }
\label{tab:ZSC_full_results}
\end{table*}

\begin{table*}[]
\centering
\renewcommand{\arraystretch}{1.2}
\begin{tabular}{c|c|cccccc}
 & \multirow{2}{*}{\textbf{Model}} & \multicolumn{6}{c}{\textbf{Language Pair and Direction}} \\ \cline{3-8} 
 &  & \textbf{lb$\leftarrow$de} & \textbf{lb$\rightarrow$de} & \textbf{lb$\leftarrow$en} & \textbf{lb$\rightarrow$en} & \textbf{lb$\leftarrow$nl} & \textbf{lb$\rightarrow$nl} \\ \hline
\multirow{4}{*}{\rotatebox[origin=c]{90}{Proprietary}} & \textbf{Cohere/embed-multilingual-light-v3.0} & 43.64 & 54.91 & 46.58 & 54.11 & 44.33 & 57.04 \\
 & \textbf{Cohere/embed-multilingual-v3.0} & 52.60 & 58.38 & 57.88 & 65.75 & 52.23 & 69.42 \\
 & \textbf{OpenAI/text-embedding-3-large} & 56.36 & 51.73 & 53.08 & 58.22 & 59.11 & 57.73 \\
 & \textbf{OpenAI/text-embedding-3-small} & 46.24 & 37.57 & 34.25 & 38.36 & 42.61 & 36.77 \\ \hline
\multirow{7}{*}{\rotatebox[origin=c]{90}{Open-Source}} &  \textbf{mBERT (CLS)} & 25.14 & 25.43 & 11.64 & 18.15 & 25.43 & 27.84 \\
 & \textbf{mBERT (MEAN)} & 36.71 & 26.88 & 22.60 & 22.26 & 34.02 & 28.18 \\
 & \textbf{LuxemBERT (CLS)} & 47.40 & 54.05 & 6.85 & 7.53 & 9.28 & 6.53 \\
 & \textbf{LuxemBERT (MEAN)} & 62.14 & 65.32 & 11.99 & 15.41 & 13.75 & 13.40 \\
 & \textbf{LASER} & 57.80 & 59.25 & 64.73 & 66.44 & 62.54 & 67.01 \\
 & \textbf{LaBSE} & \textbf{67.63} & 67.63 & \textbf{70.89} & \textbf{70.89} & \textbf{73.54} & 70.10 \\
 & \textbf{\tool} & 66.47 & \textbf{68.50} & \textbf{70.89} & 69.18 & 73.20 & \textbf{73.20} \\ \hline
\end{tabular}
\caption{\textbf{Bitex Mining Performance} on the Tatoeba dataset for three different language pairs. Retrieval accuracy values are provided for each language pair in both retrieval directions.}
\label{tab:tatoeba_full_results}
\end{table*}

\end{document}